\documentclass{article}
\usepackage{spconf,amsmath,graphicx}

\usepackage{breqn}
\usepackage{xcolor}
\usepackage{indentfirst}
\usepackage{amsmath}
\usepackage{cases}
\usepackage{booktabs}
\usepackage{multirow} 
\usepackage{amssymb}
\title{FAST PERSONALIZED TEXT TO IMAGE SYNTHESIS WITH ATTENTION INJECTION}
\name{
Yuxuan Zhang\textsuperscript{\rm 1}, Yiren Song\textsuperscript{\rm 1}, Jinpeng Yu\textsuperscript{\rm 2}, Han Pan\textsuperscript{\rm 1}, Zhongliang Jing$^{\dagger}$ \thanks{$^{\dagger}$Corresponding author}\textsuperscript{\rm 1}
}
\address{    
\textsuperscript{\rm 1}Shanghai Jiao Tong University, \textsuperscript{\rm 2}ShanghaiTech University\\
$\{$ zyx153, songyiren, hanpan, zljing $\}$ @sjtu.edu.cn, yujp1@shanghaitech.edu.cn
}

\begin{document}
%\ninept
%
\maketitle
\begin{abstract}
Currently, personalized image generation methods mostly require considerable time to finetune and often overfit the concept resulting in generated images that are similar to custom concepts but difficult to edit by prompts. We propose an effective and fast approach that could balance the text-image consistency and identity consistency of the generated image and reference image. Our method can generate personalized images without any fine-tuning while maintaining the inherent text-to-image generation ability of diffusion models. Given a prompt and a reference image, we merge the custom concept into generated images by manipulating cross-attention and self-attention layers of the original diffusion model to generate personalized images that match the text description. Comprehensive experiments highlight the superiority of our method.
\end{abstract}

\section{Introduction}
\label{sec:intro}
Personalized image generation focuses on extracting features from a selected reference image (concept) based on a given text prompt and injecting them into the generated high-quality image. The generated images should conform both to the textual description (text-image consistency) and the basic characteristics of the custom concept (identity consistency).

Previous personalized image generation methods can be divided into two main categories: tuning-based methods and encoder-based methods. Tuning-based methods~\cite{TI,lora,DB,customdiffusion, dreamartist,svdiff, breakascene} mainly convert a concept into a text embedding or convert the concept within the parameters of a generative model through fine-tuning. These methods often need lots of time to fine-tune and often struggle to maintain a balance between text-image consistency and identity consistency because the training of these methods tends to overfit reference images. Encoder-based methods~\cite{elite,designaencoder, domainagnostic,instantbooth, face0,taming} mainly train a mapping between the concept images and the text embeddings. These methods improve the time problem but are poor in text-image consistency and identity consistency. They often have difficulty creating images that accurately reflect the textual descriptions and are similar to the custom concepts. Although most of these encoder-based methods show great results in their paper, ~\cite{instantbooth} found that training a text embedding mapping also tends to make the model overfit to the training data, losing consistency with the text prompts. Additionally, most of these encoder-based methods are not open source which makes it hard for us to evaluate. 

Inspired by recent works ~\cite{p2p,nullTI,parmar2023zero} in image-to-image translation, which typically edits images through diffusion's attention layers, we found it is not necessary to train any additional text embedding for each concept. We propose a novel approach, which directly uses a coarse description like `woman' to represent the identity we want to customize and replace the original concept with an identity provided by the user during the image generation process.

\begin{figure}[t]
    \centering
    \centering
    \includegraphics[width=0.48\textwidth]{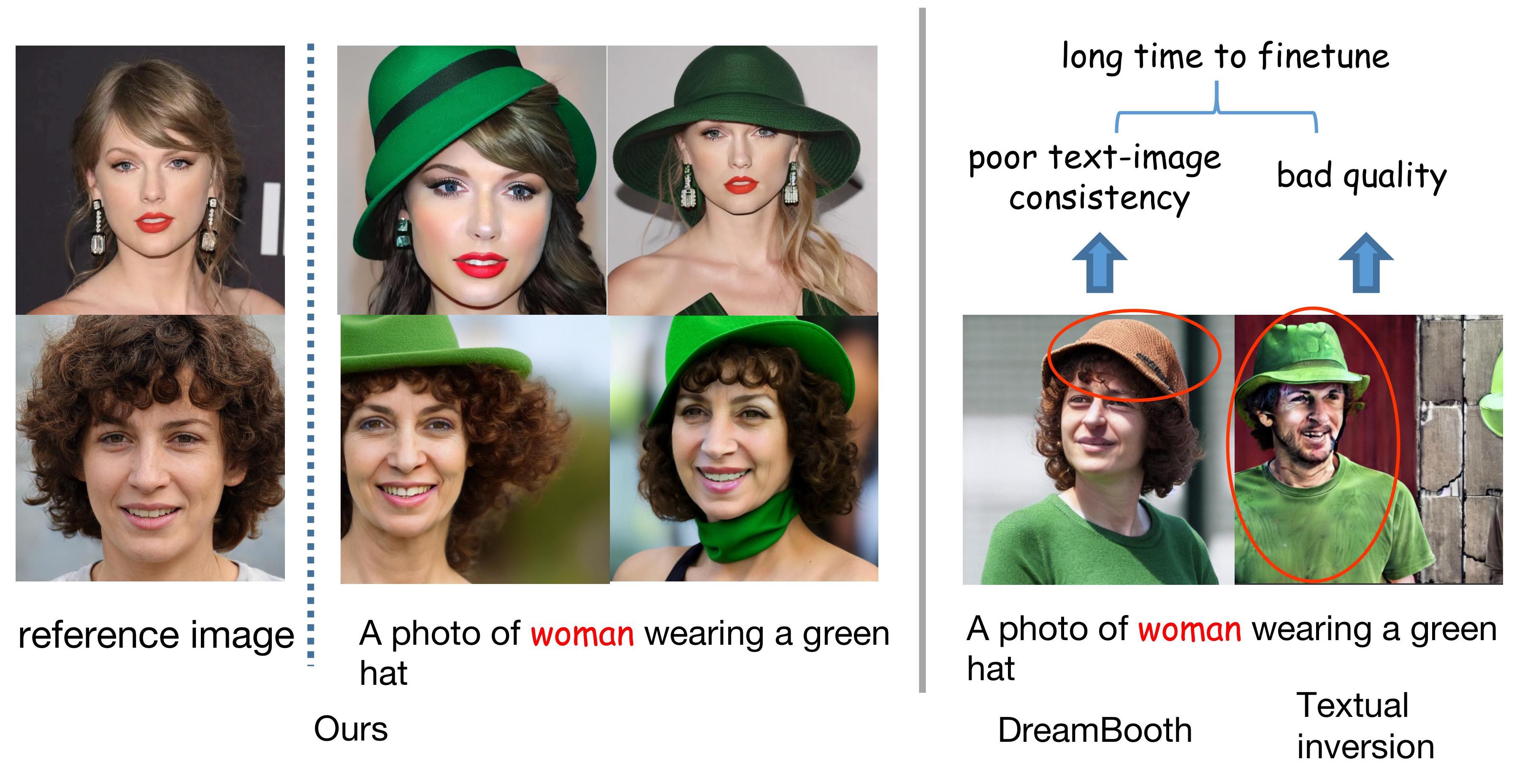}
    \vspace{-0.9cm}
    \caption{The method we proposed is a fast personalized image generation approach, which does not require any fine-tuning or optimization and only needs one image for inference. It has shown better results in terms of text-image consistency and generation quality than other methods while maintaining identity consistency.}
    \vspace{-0.3cm}
    \label{image1}
\end{figure}

We use a dual-UNet structure and propose two attention injection manipulations called masked self-attention injection and cross-attention direct detail injection, which are used to generate personalized images. As illustrated in Fig.\ref{image1}, these two attention manipulations maintain the original synthesis abilities of the pre-trained text-to-image model that improve the text-image consistency and the generated image's quality while also assuring identity consistency between the concept and the generated image.

Our key contributions are as follows: 1) We introduce a novel approach inspired by image-to-image translation works, which employs attention injection for concept customization rather than training additional text embeddings. 2) Our approach preserves the original text-to-image synthesis capabilities of the generative model which assures text-image consistency, generative quality, and identity consistency. The efficacy of our method is proved by extensive experiments. 3) Our fast method only requires one image for inference and does not require optimization or fine-tuning for each concept.

\vspace{-0.2cm}

\section{Method}
\label{sec:format}
\subsection{Overall Framework}

\begin{figure}[t]
    \centering
    \includegraphics[width=0.55\textwidth]{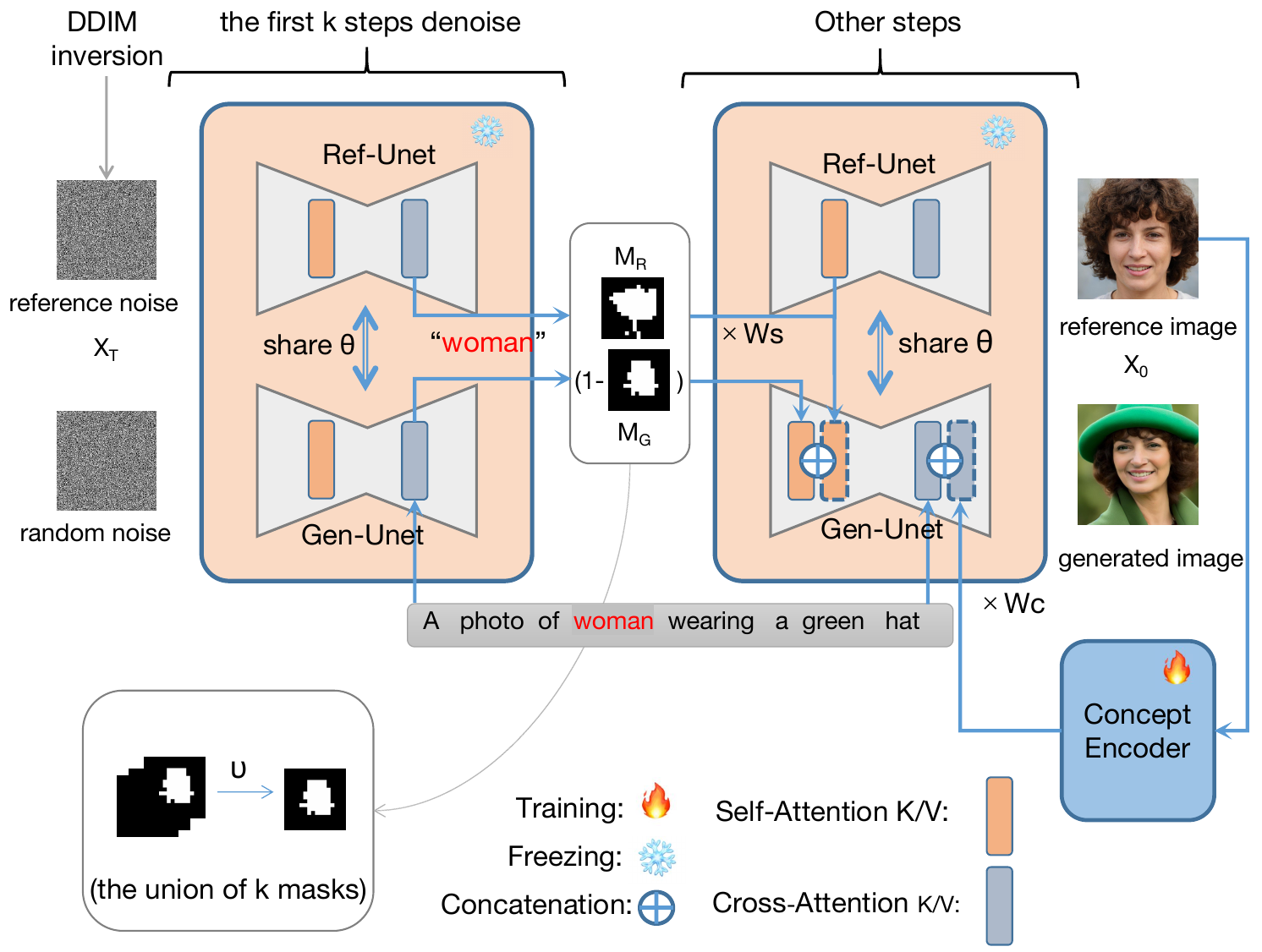}
    \vspace{-0.6cm}
    \caption{Overall schematics of our method.}
    \vspace{-0.5cm}
    \label{method}
\end{figure}

Our method is based on a pre-trained text-to-image diffusion model, with the overall framework shown in Fig\ref{method}. Given a reference image and a textual prompt, our method replaces the original concept during the generation process. Our method uses DDIM inversion to encode the reference image to obtain a series of latent noise features $\left\{X_0, X_1, X_2, \cdots  X_T\right\}$. Then, a noise is randomly sampled and this random noise as well as the previously obtained noise latents from the reference image are input into the dual U-Net structure. During each step of denoising the random noise, both masked self-attention injection and cross-attention direct detail injection are used to generate personalized images.

\subsection{Masked Self-attention injection}

We concatenate the key and value features of self-attention from Ref-Unet and Gen-Unet along the spatial dimension, which not only retains the original Unet's ability to generate images from prompt but also injects the identity information from the Ref-Unet into the Gen-Unet. Additionally, we use a cross-attention map to filter and enhance the self-attention's key and value features from Ref-Unet.

Our masked self-attention injection can be expressed as:

\begin{equation}
\mathbf{f}_G^{\prime}= \begin{cases}\mathbf{f}_G \oplus \mathbf{f}_R & \text { if } t \leq (T-k)  \\
\mathbf{f}_G & \text { otherwise }\end{cases}
\end{equation}

\vspace{-0.5cm}

\begin{equation}
\text {then, }Q=W^Q\left(\mathbf{f}_G\right), K=W^K\left(\mathbf{f}_G^{\prime}\right),V=W^V\left(\mathbf{f}_G^{\prime}\right),
\end{equation}
\vspace{-0.5cm}

\begin{dmath}
\operatorname{Attention}(Q, K, V)=W_S\left(M_R \operatorname{Softmax}\left(\frac{Q K^T}{\sqrt{d}}\right)\oplus \left(1-M_G\right)\operatorname{Softmax}\left(\frac{Q K^T}{\sqrt{d}}\right)\right) V^T
\end{dmath}
\vspace{-0.3cm}
Where $f_R$ and $f_G$ represent the spatial features of Ref-Unet and Gen-Unet, respectively. Operater $ \oplus $ represents the concatenation operation. $W^Q, W^K, W^V$ represent the projection layer, $M_G$ and $M_R$ represent the masks obtained by thresholding the normalized cross-attention map of the corresponding coarse description's token. $t$ refers to the current step of the denoise process. $T$ is the total timesteps of the denoise process which is set to 50. $k$ refers to the timesteps of mask initialization which is a constant 2. During concatenation, $W_S$ is a weight to adjust the output size of the softmax and it influences identity consistency as shown in Fig.\ref{ab}. The operation on the mask can be expressed as:
\vspace{-0.3cm}
\begin{equation}
\operatorname{normalize}(\mathbf{X})=\frac{\mathbf{X}-\min _{h, w}(\mathbf{X})}{\max _{h, w}(\mathbf{X})-\min _{h, w}(\mathbf{X})} \
\end{equation}
\vspace{-0.3cm}
\begin{equation}
M =\operatorname{normalize}(\operatorname{sigmoid}(\operatorname{normalize}(\mathcal{A}^{K})-0.5)) \\
\end{equation}

Where $\mathcal{A}^{K}$ means the cross-attention map corresponding to the k-th token.

\begin{figure}[htb]
    \centering
    \vspace{-0.3cm}
    \includegraphics[width=0.47\textwidth]{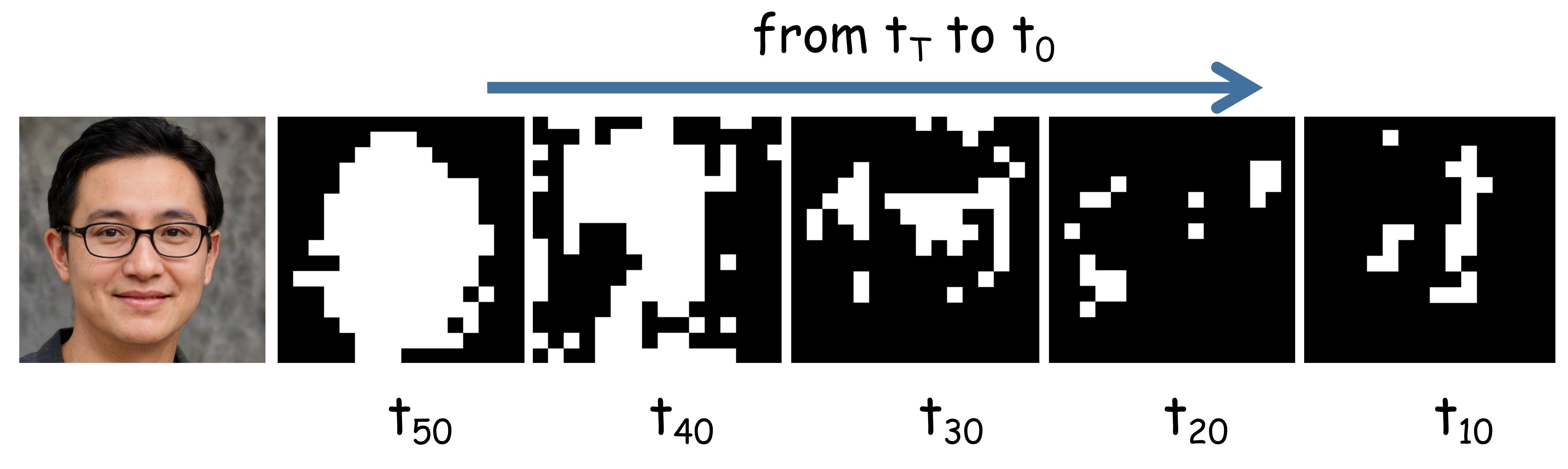}
    \vspace{-0.3cm}
    \caption{The visualization of the cross-attention map after normalization.}
    \vspace{-0.2cm}
    \label{mask}
\end{figure}

As shown in Fig.\ref{mask}, the mask obtained in the early stage of denoise is more faithful to the true position in the image of the concept. Therefore, we use the union of the masks of the first two steps in order to obtain a more reasonable mask.

\begin{figure}[h]
    \centering
    \vspace{-0.3cm}
    \includegraphics[width=0.49\textwidth]{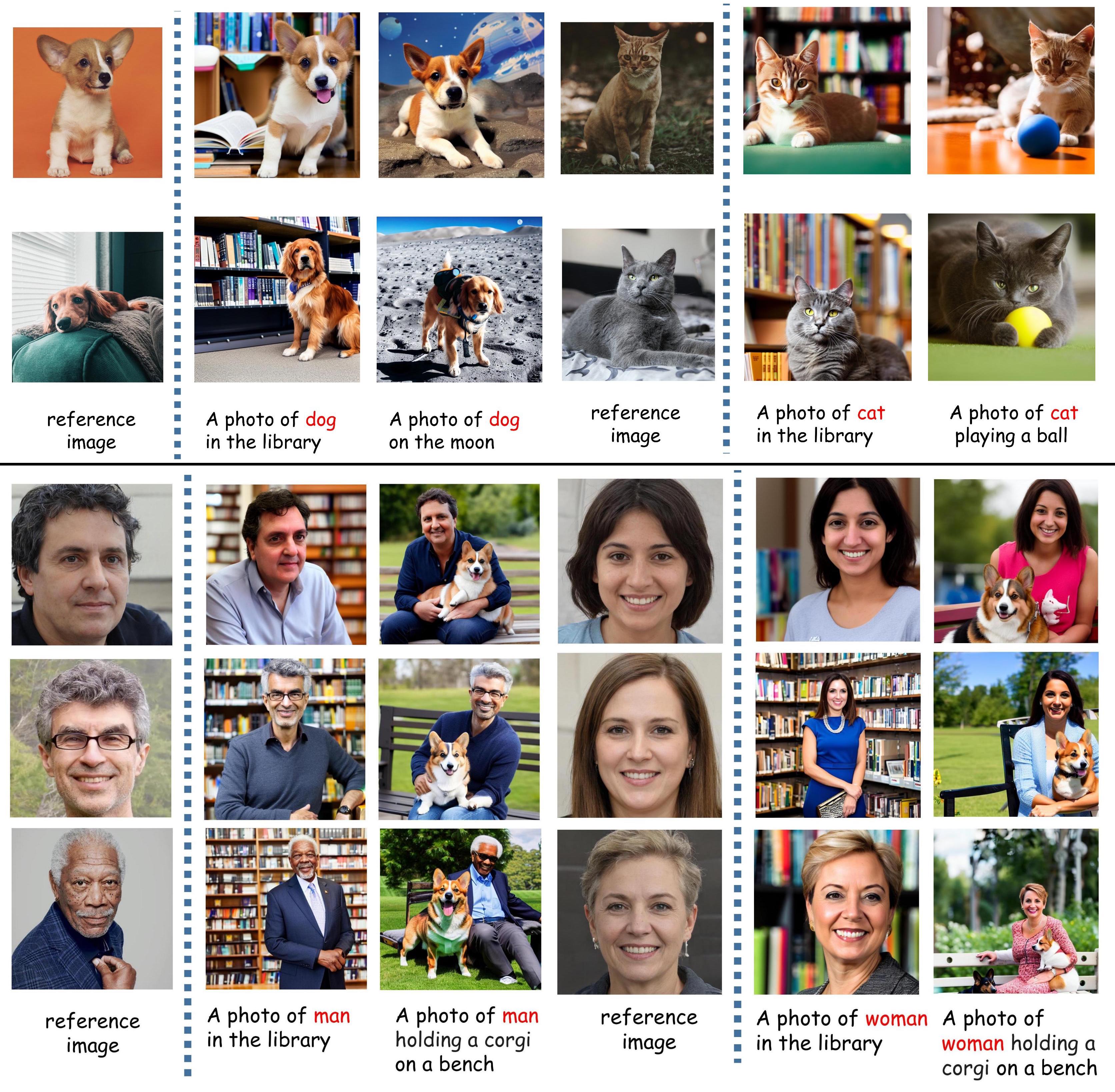}
        \vspace{-0.5cm}
    \caption{Our method results on different domains.}
        \vspace{-0.2cm}
    \label{res}
\end{figure}

\subsection{Cross-attention direct detail injection}
We directly predict the cross-attention key and value features in the Gen-Unet using an encoder and then concatenate them with the cross-attention key and value features.

The structure of cross-attention injection can be simply represented as:
\vspace{-0.3cm}
\begin{equation}
\label{eq6}
\mathbf{f}=\mathrm{MLP}\left( \mathrm{Encoder}\left(I_R\right)\right)
\end{equation}
\vspace{-0.3cm}
\begin{equation}
\mathbf{f}^{\prime}= \begin{cases}\mathbf{f}_{prompt} \oplus \mathbf{f} & \text { if } t \leq (T-k) \\
\mathbf{f}_{prompt} & \text { otherwise }\end{cases}
\end{equation}

\vspace{-0.3cm}

\begin{equation}
\text {then, }Q=W^Q\left(\mathbf{f}_G\right), K=W^K\left(\mathbf{f}^{\prime}\right), V=W^V\left(\mathbf{f}^{\prime}\right)
\end{equation}

\vspace{-0.6cm}

\begin{equation}
\operatorname{Attention}(Q, K, V) = 
\left(W_C\operatorname{Softmax}\left(\frac{Q K^T}{\sqrt{d}}\right) \right)V^T
\end{equation}
where $f_G$ is the spatial feature of Gen-Unet, $f_{\text{prompt}}$ is the output of the CLIP text encoder, and $I_R$ represents the reference image. $W^Q, W^K, W^V$ represent the projection layer. $W_S$ is a weight to adjust the output size of the softmax and it also influences identity consistency as shown in Fig.\ref{ab}. 

The structure of our encoder can be represented as Equation(6), where the Encoder is the image encoder from CLIP~\cite{clip}, extracting the rich detailed features embeddings (257$\times$768) from the penultimate layer, and using an MLP as the projection layer to project the rich feature embeddings into the cross-attention key and value features. 

During training, we train a concept encoder to reconstruct every image in the dataset of a specific domain. Our loss function to optimize the model is formulated as:
\begin{equation}
\mathcal{L}=\mathbb{E}_{z, t, c, X_s, \eta \in \mathcal{N}(0,1)}\left[\left\|\eta-\eta_\theta\left(z_t, t, f, X_s\right)\right\|_2^2\right]
\end{equation}
where $z_t$ is the latent noisy image at time step $t$ obtained from the training images, $\eta$ is the latent noise grand-truth. $X_s$ is the set of training images, and $\eta_\theta$ is the noise prediction model with parameters $\theta$. $f$ is the concept encoder's output. Both the projection layer and encoder are updated.

During inference, we simply input reference images into the concept encoder and concatenate the output embedding with the cross-attention key and value features from the original text encoder.
\vspace{-0.5cm}
\section{EXPERIMENTAL RESULTS}
\begin{figure}[t]
    \centering
    \vspace{-0.3cm}
    \includegraphics[width=0.49\textwidth]{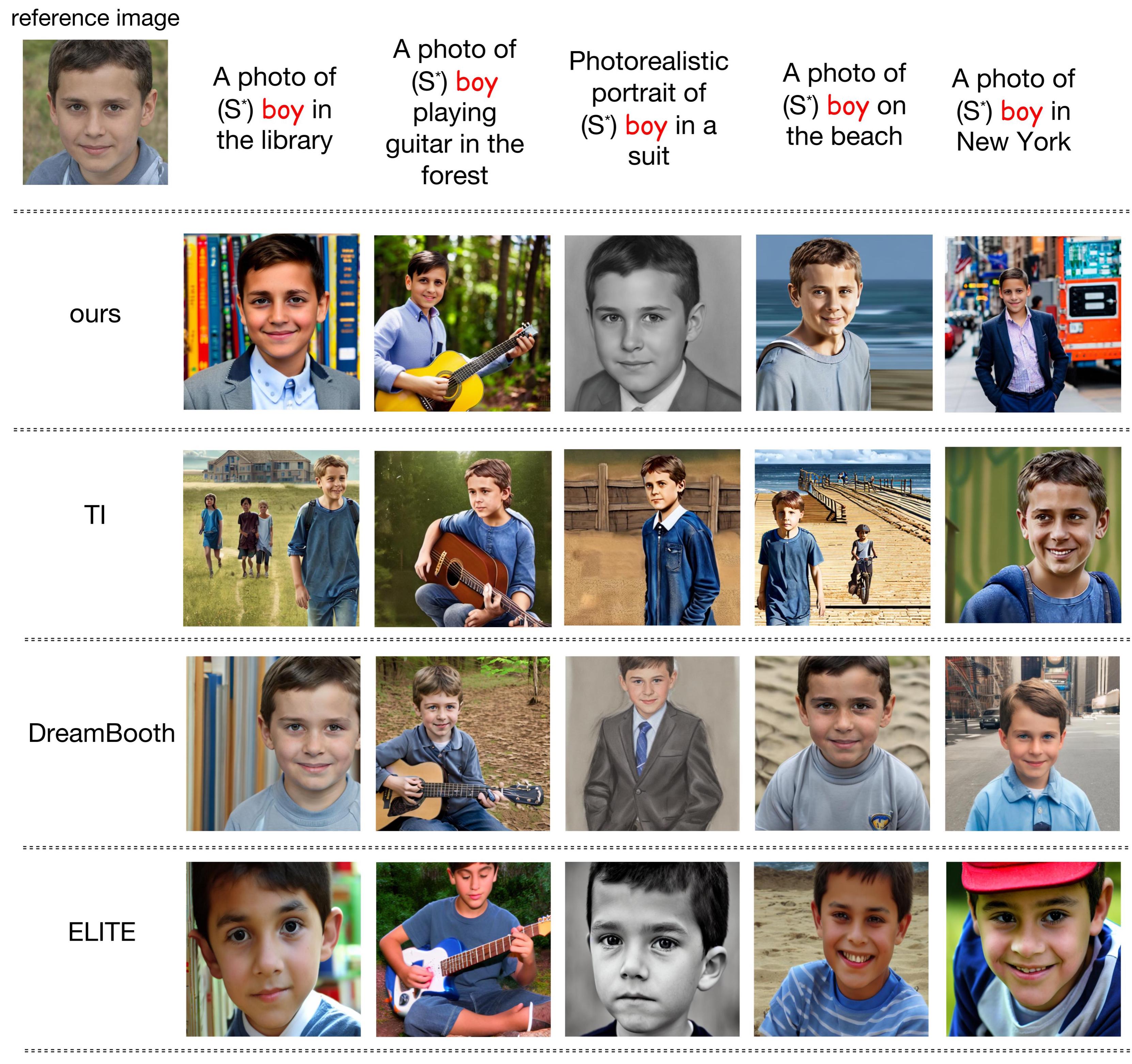}
    \caption{The visual comparison with other methods.}
    \vspace{-0.3cm}
    \label{compare}
\end{figure}
\subsection{Comparation and Evaluation}
The pre-trained text-to-image model we used is Stable Diffusion V1-5. During training, our encoder is trained on FFHQ~\cite{FFHQ} and AFHQ~\cite{AFHQ} for 100k iterations on four A100s with a batch size of 2 and a learning rate of 1e-5. Taking the human face domain as an example, we compare our method with TI, DB, and ELITE, which are the three most popular algorithms at present. 
\vspace{-0.5cm}
\subsubsection{Metrics}
\textbf{1. Text-Image Consistency:} We use CLIPScore~\cite{clipscore} to calculate the consistency between the generated image and the text prompts. \textbf{2. Identity Consistency:} Take the face domain as an example. To measure identity consistency, we use a face detector~\cite{dlib} to detect faces in both the generated images and the reference images. Then we extract embeddings from each detected face using an off-the-shelf method~\cite{facenet} and calculate the average embedding distance for each pair of faces. 
\textbf{3. Generative Quality:} We score the mean aesthetics of the generated images using an aesthetic predictor~\cite{laion5b} to measure the generative quality. 
\textbf{4. Time:} Since time directly affects our actual application, we also take the time cost as a comparison indicator.

\vspace{-0.3cm}
\subsubsection{Qualitative Results}
The comparison results for personalized face generation can be seen in Fig.\ref{compare}. Generally, TI tends to produce blurred backgrounds or even images that fail to match the description text, which also is a problem occasionally encountered in ELITE's results. DB's identity consistency when generating a big face is fairly good, although, as shown in the second column, it falls short in maintaining identity consistency when generating smaller faces. Our method shows the best performance in generative quality and maintaining consistency between the image and text while maintaining identity consistency. More results from our method can be seen in Fig.\ref{res} and supplementary documents.

\vspace{-0.3cm}
\subsubsection{Quantitative Evaluation}
\vspace{-0.3cm}
\setlength{\tabcolsep}{1mm}{
\begin{table}[htb]
\begin{footnotesize}
\centering
\begin{tabular}{*{2} ccccc}%5个c代表该表一共5列
\toprule
 & \multirow{2}*{\small Methods} & \small T-I & \small Identity & \small Generative & \multirow{2}*{\small Time(s)$\downarrow$} \\
&  & \small Consistency$\uparrow$ & \small Consistency$\uparrow$ & \small Quality$\uparrow$ &  \\
 
\midrule%第二道横线 
 &TI &0.2332 & 1.4261 & 6.4249 & $>$1000\\
 &DB &0.3032 & \textcolor{red}{1.5271} & 6.6285 & $>$500\\
 &ELITE &0.2109 & 1.2172 & 6.0725 & $<$50\\
 &Ours &\textcolor{red}{0.3526} &1.4251 & \textcolor{red}{6.6489} & \textcolor{red}{$<$10}\\
\bottomrule%第三道横线
\end{tabular}
\caption{\textrm{Quantitative comparison of different methods.}}%标题
\vspace{-0.4cm}
\end{footnotesize}
\end{table}
}

\setlength{\tabcolsep}{0.7mm}{
\begin{table}[htb]
\begin{footnotesize}
\centering
\begin{tabular}{*{2} cccc}
\toprule
 & \multirow{2}*{\small Methods} & \small T-I & \small Identity & \small Generative \\
&  & \small Consistency$\uparrow$ & \small Consistency$\uparrow$ & \small Quality$\uparrow$ \\
 
\midrule%第二道横线 
 &w.o. injection &\textcolor{red}{0.3814} & 0.8392 & \textcolor{red}{6.6586} \\
 &only cross-attn injection &0.3227 & 1.3298 & 6.0573\\
 &only self-attn injection  &0.3164 &1.3103 & 6.4166\\
 &full & 0.3526 & \textcolor{red}{1.4251} &6.6489 \\
\bottomrule%第三道横线
\end{tabular}
\caption{\textrm{Quantitative comparison of different components and settings.}}%标题
\vspace{-0.3cm}
\end{footnotesize}
\end{table}
}
The quantitative comparison with other methods is shown in Table 1. Our method is close to DB in identity consistency, and it achieves the best in text-image consistency. Since our method does not require fine-tuning or training for any concept, our speed advantage is significant compared to other algorithms. Furthermore, the aesthetic score of the images we generate leads these algorithms. 

The quantitative comparison of different components and settings is shown in Table 2. When all components work together, the best identity consistency is achieved. Furthermore, the text-image consistency and aesthetic scores of our method are nearly equivalent to those of the original SD without any attention injection. 

Better T-I consistency, generative quality, and high identity consistency prove that our method does not destroy the original text-to-image generative power of the diffusion model and makes a better text-image consistency and generative quality than the methods above while maintaining identity consistency. 

\subsection{Ablation Study}
\begin{figure}[h]
    \centering
    \vspace{-0.4cm}
    \includegraphics[width=0.49\textwidth]{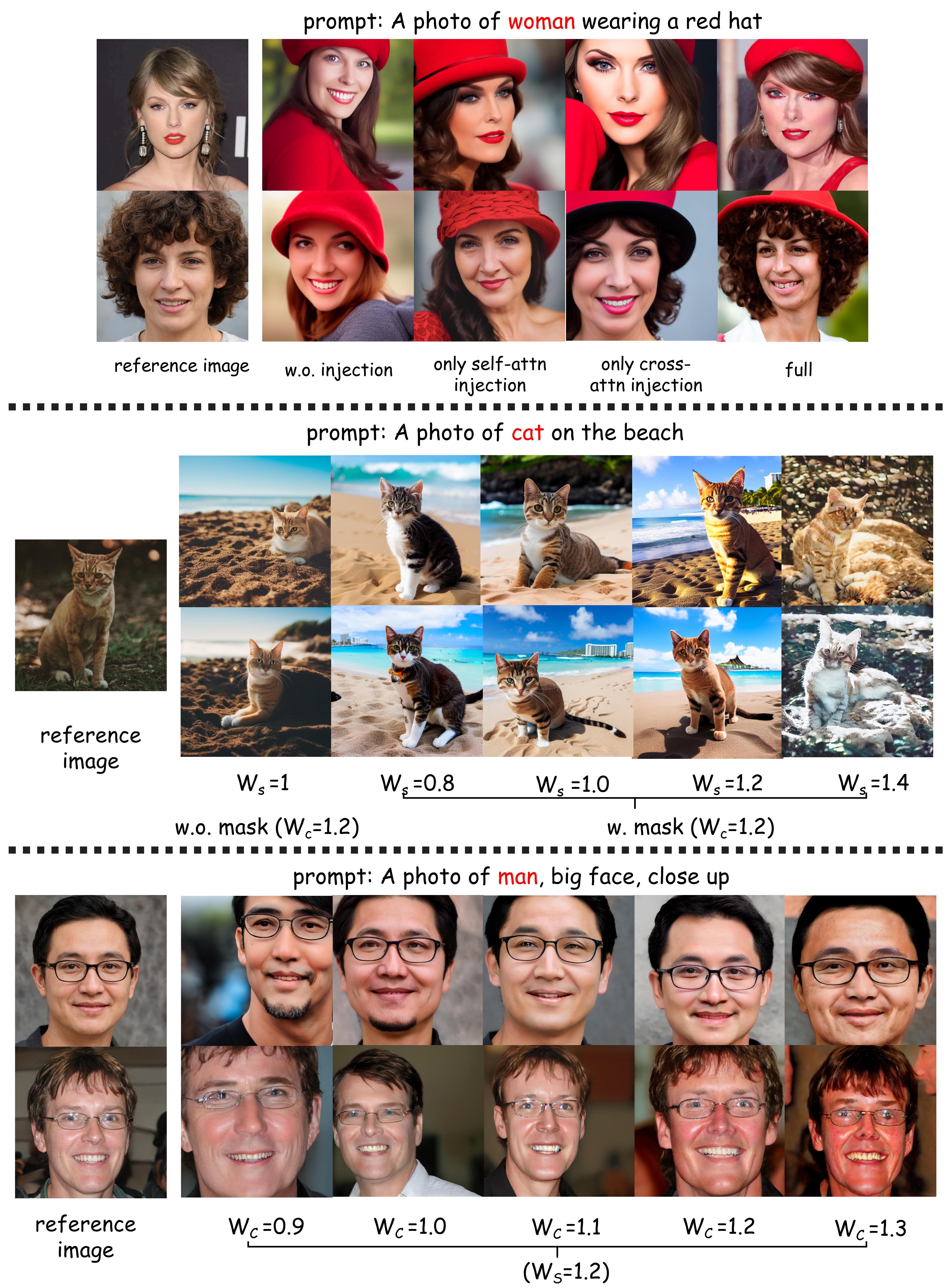}
    \vspace{-0.6cm}
    \caption{The comparison of different ablation settings.}
    \vspace{-0.6cm}
    \label{ab}
\end{figure}

We conduct a thorough ablation study across various components and settings. As shown in Fig.\ref{ab}, when only self-attention injection or cross-attention injection, the generated image has a closer resemblance to the reference image but lacks facial details. The full injections have more faithful identity preservation. The second row of Fig.\ref{ab} demonstrates that without a mask, beach images generated with self-attention injection result in black mudflats that match the color tone of the reference image instead of smooth sandy beaches. Additionally, a larger $W_S$ value produces a more similar identity, but exceeding a certain threshold leads to image deterioration. The third row of Fig.\ref{ab} demonstrates that increasing the value of $W_C$ can heighten the level of identity consistency. Too small will generate an image that is not similar to the reference image, and too large will also cause image distortion.

\section{Conclusion}
We have proposed a new fast personalized image generation method. Our method improves the generative quality and text-image consistency of personalized image generation methods while ensuring identity consistency. Additionally, it does not require fine-tuning or training for each concept. Its key idea is to use Ref-Unet and a domain-specific concept encoder to merge the concept into the Gen-Unet by manipulating self-attention and cross-attention layers during generation. This rapid method, with its great generative quality and superior balance between text-image consistency and identity consistency, advances the field of personalized image generation, providing increased flexibility in applications.

\vfill\pagebreak

\bibliographystyle{IEEEtran}

\end{document}